# SSD-FASTER NET: A HYBRID NETWORK FOR INDUSTRIAL DEFECT INSPECTION

*Jingyao Wang, Naigong Yu*

Beijing Key Lab of the Computational Intelligence and Intelligent System, Beijing University of Technology, China

## ABSTRACT

The quality of industrial components is critical to the production of special equipment such as robots. Defect inspection of these components is an efficient way to ensure quality. In this paper, we propose a hybrid network, SSD-Faster Net, for industrial defect inspection of rails, insulators, commutators etc. SSD-Faster Net is a two-stage network, including SSD for quickly locating defective blocks, and an improved Faster R-CNN for defect segmentation. For the former, we propose a novel slice localization mechanism to help SSD scan quickly. The second stage is based on improved Faster R-CNN, using FPN, deformable kernel(DK) to enhance representation ability. It fuses multi-scale information, and self-adapts the receptive field. We also propose a novel loss function and use ROI Align to improve accuracy. Experiments show that our SSD-Faster Net achieves an average accuracy of 84.03%, which is 13.42% higher than the nearest competitor based on Faster R-CNN, 4.14% better than GAN-based methods, more than 10% higher than that of DNN-based detectors. And the computing speed is improved by nearly 7%, which proves its robustness and superior performance.

## 1. INTRODUCTION

The quality of components is crucial for the assembly of industrial production lines or special robots. Among them, steel pipes and other products are also widely used in high-pressure and high-risk scenarios such as petroleum, chemical, and natural gas. If such devices are defective or broken, there will be serious and irreversible consequences. With the development of intelligence and automation in China, the demand for high-precision and high-quality components is also increasing. However, due to the production environment and storage conditions, various problems will occur on the surface of the device. Such as scratches, scaling, surface cracks, detail defects, etc., resulting in reduced corrosion resistance, wear resistance and fatigue strength etc. For different electronic and metal components, the defects change randomly. Figure 1 depicts examples of data used in this paper. The rapid realization of defect inspection and evaluation has become a hot topic. We propose a two-stage hybrid network that can quickly locate defects and complete segmentation. This method enables the detection of insulators, commutators, rails etc., and can be activated or deactivated at any time as needed.

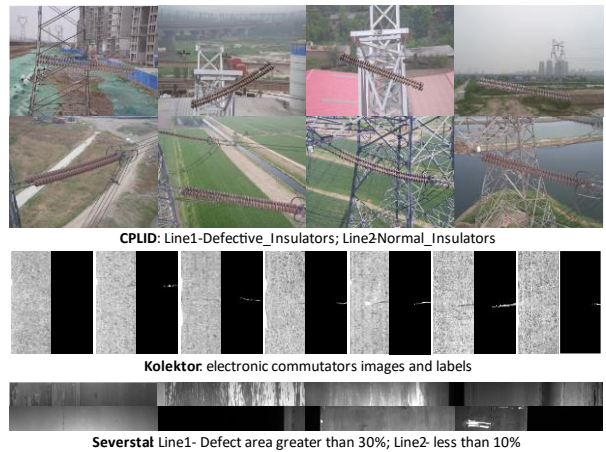

**Fig. 1**: Samples of datasets used for SSD-Faster Net

The traditional methods of industrial defect inspection include manual inspection [1] and X-ray inspection [2]. The former mainly judges whether the industrial components has quality problems such as cracks and scratches through human eyes. The latter is the main method of traditional industrial non-destructive testing, and its results are often used as an important basis for the quality evaluation of metal weld defects. This method effectively inspects internal defects through X-ray scanning devices, but still requires manual participation [3]. Traditional methods rely on manual labor, which has low efficiency and large individual differences in accuracy.

With the development of deep learning, industrial defect inspection has become more intelligent. In order to meet the needs of contemporary "smart industrialization", computer vision-based industrial detection has gradually replaced traditional manual-based methods. In classic machine vision, learning-based classifiers such as decision tree [4], SVM [5], KNN [6], and texture-based methods such as GLCM [7], AELTP [8], AECLBP [9], etc., use manual processing feature for defect detection. These algorithms mainly based on filter [10], wavelet transform [11], morphological operation [12] etc. to manually create the features of annotation.

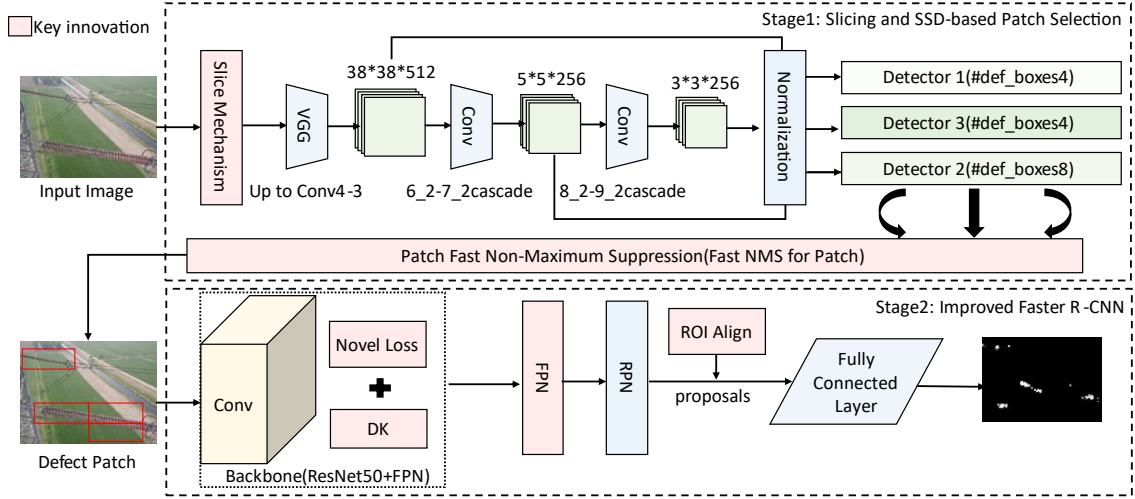

**Fig. 2**: The framework of SSD-Faster Net.

Although these algorithms are accurate for small data processing, the manual nature makes them inflexible and unsuitable for long industrial production cycles. Research for industrial defect detection has gradually focused on deep learning. Deep learning-based algorithms such as OSTU + MSVM-rbf (Multi-class Support Vector Machine) [13], GAN [14], CFM [15], TGRNet [16] have achieved good results. But there are still shortcomings: poor transferability, different types of defects make it hard to perform multiple classifications; bad real-time performance, the checking time is too long and it is difficult to apply to the industrial field.

Therefore, we propose a hybrid network to solve the above problems and further improve the accuracy. Our contributions are as follows:

- We propose a novel hybrid network called SSD-Faster Net. The whole framework is divided into two stages, which combines SSD [17] and Faster R-CNN [18] model. The former completes the initial screening of the input and quickly locates the defect patch. The latter segment the selected patch and extract defective pixels. Experiments show that the network achieves an average accuracy improvement of 6.94% on the three defect datasets, and surpasses the calculation speed by nearly 7%.
- We propose the SSD-based fast slice localization mechanism. By dividing the input data, the algorithm helps the SSD to quickly locate the patches where defects exist in the first stage.
- We improve the Faster R-CNN of the hybrid network. We use FPN (feature pyramid) with a deformable kernel(DK) to enhance model representation. In addition, ROI Align replaces ROI Pooling to further improve the model accuracy. We also design a novel loss function to get better performance.
- We validate the robustness of the model on three different datasets, rather than focusing on just one defect type. At the same time, we have designed a modular structure that can be activated or deactivated as needed. This architecture also facilitates the testing and analysis of the model on a variety of datasets.

The rest of this article is organized as follows. In Section 2, we describe the proposed two-stage hybrid network in detail. Section 3 introduces the relevant datasets, detailed design and results of the experiments. The conclusion part is in the fourth section.

## 2. PROPOSED HYBRID NETWORK

Our SSD-Faster Net is a multi-stage hybrid network. It performs industrial defect inspection by segmenting and positioning the input RGB image. Different from the target recognition in previous study, SSD-Faster Net performs pixel-level segmentation. It introduces a fast localization mechanism and a deformable kernel to achieve precise localization of defects. Compared with baseline methods, SSD-Faster Net is more efficient and pay more attention to the sensitive features of large-scale datasets.

SSD-Faster Net slices the input regardless of the image quality used for the initial computation. Based on the sliced patch, our network utilizes SSD with improved Faster RCNN for fast segmentation. We adopt a scanning method that combines "global" and "local" features to make the network more robustness. SSD-Faster Net has better performance on industrial components no matter electronic commutators or metal devices. The modular structure and simple interface enable each part to be activated as needed, which make the entire network be more flexible. The specific structure of SSD-Faster Net is shown in Figure 2.

### 2.1. Architecture Overview

Figure 2 shows the architecture of SSD-Faster Net. The hybrid network mainly consists of two main parts. The first stage takes RGB images as input and uses slice localization to segment it: the entire image is divided into multiple patches. The number of patches is designed in advance according to the characteristics such as the size of the input image. The SSD method scans the segmented patch to locate defective blocks, and the result will be sent to the second stage for further calculation and segmentation. This rapid initial screening narrows the target range and greatly increases the speed of defect detection.

The second stage performs accurate segmentation and pixel-level defect inspection. This stage is mainly based on improved Faster RCNN. The network contains four submodules. In addition to replacing the backbone and other links of the original network according to the experimental results, we also introduce operations such as FPN [19], deformable kernel [20], and ROI align [21]. These innovations enable SSD-Faster Net to adapt to the effective receptive field and improve representation ability. At the same time, this operation of integrating multi-scale design and eliminating floating point errors greatly improves the accuracy. After constant iterations, we finally achieve superior results.

**2.2. Stage1: Slicing and SSD-based Patch Selection**

The first stage of SSD-Faster Net mainly consists of two parts: multi-scale slicing and SSD-based defect screening. Firstly, we need to segment the incoming RGB image. Referring to the anchors design rules of the RPN [18], a total of three areas {64,128,256}{64,128,256} and three proportions of patch blocks {1:1,1:2,2:1}{1:1,1:2,2:1} are selected in the slices in this section. Before this, the network will convert the image into a fixed size through operations such as grayscale, geometric transformation, and resize. After slicing, the original RGB image will be processed into $n$ patches that remove redundant information.

Next, we scan the patch based on SSD. The target is filter out patch blocks that contain defects. In addition, the raw data of some industrial components contains much background information, such as electronic commutators. This step can greatly reduce the segmentation time of the next stage by finding the patch blocks containing the objects to be detected. Different from the 6 different sizes of feature maps of the original SSD, only three Default boxes are used in SSD-Faster Net. The network structure is shown in Figure 2.

$$J(a,b) = \frac{|A \cap B|}{|A \cup B|} \in [0,1] \quad (1)$$

$J(a,b)$ is the Jaccard coefficient, which is used to calculate the similarity between the screened patch and the real defective block. $A$ and $B$ are the corresponding finite sets, respectively. The overall objective loss function is the weighted sum of the localization loss and the confidence loss:

$$L(x,c,l,g) = \frac{1}{N}(L_{conf}(x,c) + \alpha L_{loc}(x,l,g)) \quad (2)$$

$L_{conf}$ is the localization loss, and $L_{loc}$ is the confidence loss. The former is a typical smooth $L1$ loss:

$$L_{loc}(x,l,g) = \sum_{i \in Pos}^{N} \sum_{m \in (x,y,w,h)} x_{ij}^k L_1(l_i^m - \hat{g}_j^m) \quad (3)$$

$l_i^m$ and $\hat{g}_j^m$ are the predicted patch and the ground truth. $(x, y)$ in this formula means center coordinates of default box, while $(w, h)$ is the length and width of the box. The latter is the Softmax loss on the multi-class confidence ($c$):

$$L_{conf}(x,c) = -\sum_{i \in Pos}^{N} x_{ij}^P \log(\hat{c}_i^P) - \sum_{i \in Neg} \log(\hat{c}_i^0) \quad (4)$$

$$\hat{c}_i^P = \frac{\exp(c_i^P)}{\sum_P \exp(c_i^P)} \quad (5)$$

$\hat{c}_i^P$ is the predicted probability of the $i$-th search box corresponding to the category $P$ (0 means background). The parameters $i$: patch number; $j$: real defect block number. The first half of the formula 4 is the positive sample (*Pos*) - defect category, and the latter is the negative sample (*Neg*) - background.

**2.3. Stage2: Improved Faster R-CNN**

In this stage, we improve Faster RCNN network to further segment the filtered patches, and get defect results.

Faster RCNN uses ResNet50 and ResNet101 networks with controllable parameters as the backbone. The feature map obtained by the residual network will be continuously under-sampled for three scale feature fusion and shared into the RPN network and Roi Pooling. Next, the network predicts the target through fully connected layers.

Considering the sensitivity of industrial defects, we improve Faster RCNN in four aspects. Because the resolution of some of the selected datasets is too large (1920*1080) but some defect blocks are small, it is difficult to distinguish the defect features even if the input is divided into patches. To obtain clearer high-order features, we introduce Feature Pyramid (FPN). We set six scales for feature extraction, and introduce an extra scale as a bias to improve the representation ability. The positions of the candidate feature boxes are scattered to each layer of the feature pyramid, which increases the feature mapping resolution of small objects.

In order to further improve the adaptive ability and robustness, we add a deformable convolution kernel (DK) to SSD-Faster Net. This novel mode samples a new convolution from the original round every time a convolution operation is performed, realizing an adaptive receptive field. In addition, to further eliminate the float error brought by ROI Pooling, we replace it with ROI Align. This bilinear interpolation method effectively avoids the loss of accuracy caused the fully connected layer.

We also update the loss function by designing patch loss. Thanks to the SSD-based fast localization, the huge computational overhead brought by NMS pixel-by-pixel is discarded. SSD-Faster Net will directly calculate the pixel confidence of the features segmented in the defective patch. The loss function uses the average binary cross-entropy loss to calculate the defects in each patch, and finally completes the semantic segmentation.

$$L = L_{cls} + L_{loc} + L_{pat} \quad (6)$$

$L_{cls}$: Category loss; $L_{loc}$: Position loss; $L_{pat}$: Average of defect confidences for each pixel within a patch.

## 3. EXPERIMENTS AND RESULTS

In this section, we first introduce the dataset used for experiments, which includes metal and electronic components. The experiments and results will be introduced in 3.2.

### 3.1. Datasets

SSD-Faster Net has designed flexible interfaces to ensure the defect detection on various industrial components. Some samples are shown in Figure 1.

**CPLID** [22] provides images of normal and synthetic defective insulators. It contains 600 normal images and 248 with defects. Some data are generated and segmented using algorithms such as TVSeg [22] and U-Net [22].

**KolektorSDD** [23] contains 50 electronic commutators, each with 8 images and labels for semantic segmentation. It contains a total of 52 images with visible defects, and 347 images used as defect-free negative examples (500*1240).

**Severstal** [24] is a steel strip dataset, which provides four types of strip surface defects. There are 12,568 images for training and 5,506 for testing. Image size is 1600*256.

### 3.2. Experimental Setup and Results

Table 1 shows the analysis results on three datasets. To comprehensively evaluate SSD-Faster Net, we report the ACC results of methods mentioned in Section 1, include SVM, KNN, GLCM etc. We take deep-learning-based GAN, CFM, TGRNet, and Faster RCNN, SSD as baselines.

**Table 1**: Defect inspection results for the three datasets(%).

| Method | CPLID | KolektorSDD | Severstal |
|---|---|---|---|
| Decision tree | 39.56 | 22.90 | 41.38 |
| SVM+HOG | 46.30 | 39.11 | 50.13 |
| KNN | 37.66 | 18.76 | 24.96 |
| GLCM+ AELTP | 31.78 | 44.51 | 27.19 |
| GAN-based | **85.73** | 69.40 | 80.52 |
| CFM-based | 62.95 | 73.91 | 71.38 |
| TGRNet | 80.01 | 62.75 | 52.41 |
| SSD | 69.33 | 51.44 | 70.56 |
| Faster RCNN | 78.64 | 67.85 | 61.33 |
| SSD-Faster Net | 83.47 | **77.31** | **87.29** |

To ensure objective results, all algorithms use the same segmentation mask. We can find that the accuracy of our SSD-Faster Net meets expectations, and the robustness is verified. It also can be clearly seen that our proposed network has certain advantages compared with methods just use single Faster RCNN or SSD.

In order to further explore the impact of different categories, defects, or image quality on performance, Table 2 shows the performance of SSD-Faster Net on 9 kinds of data. They are selected evenly in the three datasets described in Section 3.1, and the defects are gradually serious.

**Table 2**: ACC for different types of data (defect area%).

| Type | A: 10- | A: 10-30 | A: 30+ | B: 10- | B: 10-30 |
|---|---|---|---|---|---|
| ACC(%) | 71.33 | 85.90 | 89.63 | 69.52 | 79.46 |
| Type | C: 10- | C: 10-30 | C: 30+ | B: 30+ | |
| ACC(%) | 85.48 | 89.45 | 91.45 | 82.31 | |

A: CPLID[23], B: KolektorSDD[24], C: Severstal[25]. "A: 10-" Indicates that the proportion of defects in the CPLID dataset is less than 10%.

**Table 3**: The effect of training scale (KolektorSDD).

| Data-Scale | 16+104(30%) | 31+208(60%) | 52+347 (100%) |
|---|---|---|---|
| ACC-AVG(%) | 65.87 | 74.26 | 77.31 |

"16+104(30%)" contains 16 images with visible defects and 104 negatives without defects, which occupies 30% of the KolektorSDD dataset.

Compared with the insulators and commutators with complex backgrounds, the steel strip contains less interference has obtained better results. We speculate that it may be due to the missing or error in the slice screening of the first stage. But the overall results basically meet expectations. For further exploration, we also test whether the scale of datasets related to this problem.

As shown in Table 3, the larger the training scale, the higher the model accuracy. But the increase is not obvious when training scale over 60%. This shows the superiority of the structure, and future optimization needs to be considered from the network itself.

## 4. CONCLUSION

In this paper, we propose a novel hybrid network for industrial defect inspection, called SSD-Faster Net. It is a two-stage cascade structure. SSD-Faster Net first utilizes our proposed SSD-based fast localization mechanism to quickly screen out patches where defects are located. The results are further segmented based on the improved Faster RCNN in the second stage. The innovative network integrating FPN, deformable kernel, novel loss function, etc. achieves the goal of quickly and accurately locking industrial defects. In order to apply SSD-Faster Net to more devices, we also have a flexible interface design adjusted for different device characteristics. This approach enables the activation or deactivation of modules as needed. Experiments show that SSD-Faster Net has good performance in various industrial components such as rails and insulators. We will further explore the potential of the network in more complex scenarios, and conduct research on model interpretability in the future.

# 5. REFERENCES

[1] Mital, Anil, M. Govindaraju, and B. Subramani. "A comparison between manual and hybrid methods in parts inspection." *Integrated Manufacturing Systems* (1998).

[2] van Dael, Mattias, et al. "Multisensor X-ray inspection of internal defects in horticultural products." *Postharvest biology and technology* 128 (2017): 33-43.

[3] Ali, Abdul-Hannan, et al. "The reliability of defect sentencing in manual ultrasonic inspection." *NDT & E International* 51 (2012): 101-110.

[4] Hongwei, Xie, et al. "Solder joint inspection method for chip component using improved AdaBoost and decision tree." *IEEE Transactions on components, packaging and manufacturing technology* 1.12 (2011): 2018-2027.

[5] Xue-Wu, Zhang, et al. "A vision inspection system for the surface defects of strongly reflected metal based on multi-class SVM." *Expert Systems with Applications* 38.5 (2011): 5930-5939.

[6] Cetiner, Ibrahim, Ahmet Ali Var, and Halit Cetiner. "Classification of knot defect types using wavelets and KNN." *Elektronika ir elektrotechnika* 22.6 (2016): 67-72.

[7] Zhu, Dandan, et al. "Yarn-dyed fabric defect detection based on autocorrelation function and GLCM." *Autex research journal* 15.3 (2015): 226-232.

[8] Zaghdoudi, Rachid, Hamid Seridi, and Slimane Ziani. "Binary Gabor pattern (BGP) descriptor and principal component analysis (PCA) for steel surface defects classification." *2020 International Conference on Advanced Aspects of Software Engineering (ICAASE)*. IEEE, 2020.

[9] Song, Kechen, and Yunhui Yan. "A noise robust method based on completed local binary patterns for hot-rolled steel strip surface defects." *Applied Surface Science* 285 (2013): 858-864.

[10] Duch, Włodzisław. "Filter methods." *Feature Extraction*. Springer, Berlin, Heidelberg, 2006. 89-117.

[11] Zhang, Dengsheng. "Wavelet transform." *Fundamentals of Image Data Mining. Springer*, Cham, 2019. 35-44.

[12] Su, Ran, et al. "A new method for linear feature and junction enhancement in 2D images based on morphological operation, oriented anisotropic Gaussian function and Hessian information." *Pattern Recognition* 47.10 (2014): 3193-3208.

[13] Jiao, Shuhong, Xueguang Li, and Xin Lu. "An improved Ostu method for image segmentation." *2006 8th international Conference on Signal Processing*. Vol. 2. IEEE, 2006.

[14] Liu, Juhua, et al. "Multistage GAN for fabric defect detection." *IEEE Transactions on Image Processing* 29 (2019): 3388-3400.

[15] Ma, Yuansheng, et al. "Machine learning based wafer defect detection." *Design-Process-Technology Co-optimization for Manufacturability* XIII. Vol. 10962. SPIE, 2019.

[16] Xue, Wenyuan, et al. "TGRNet: A Table Graph Reconstruction Network for Table Structure Recognition." *Proceedings of the IEEE/CVF International Conference on Computer Vision*. 2021.

[17] Liu, Wei, et al. "Ssd: Single shot multibox detector." *European conference on computer vision*. Springer, Cham, 2016.

[18] Ren, Shaoqing, et al. "Faster r-cnn: Towards real-time object detection with region proposal networks." *Advances in neural information processing systems* 28 (2015).

[19] Li, Zihao, et al. "MVP-Net: multi-view FPN with position-aware attention for deep universal lesion detection." *International Conference on Medical Image Computing and Computer-Assisted Intervention*. Springer, Cham, 2019.

[20] Xu, Xuan, et al. "Deformable kernel convolutional network for video extreme super-resolution." *European Conference on Computer Vision*. Springer, Cham, 2020.

[21] He, Kaiming, et al. "Mask r-cnn." *Proceedings of the IEEE international conference on computer vision*. 2017.

[22] Insulator Data Set-Chinese Power Line Insulator Dataset (CPLID). *https://github.com/InsulatorData/InsulatorDataSet*.

[23] KolektorSDD (Kolektor Surface-Defect Dataset). *https://www.vicos.si/Downloads/KolektorSDD*.

[24] Severstal. *https://www.kaggle.com/c/severstal-steel-defect-detection/data*.